# Developing a New Autism Diagnosis Process Based on a Hybrid Deep Learning Architecture Through Analyzing Home Videos

Spencer He, Valley Christian High School, San Jose 95111
Ryan Liu, Lynbrook High School, San Jose 95129

## Abstract

Currently, every 1 in 54 children have been diagnosed with Autism Spectrum Disorder (ASD), which is 178% higher than it was in 2000. An early diagnosis and treatment can significantly increase the chances of going off the spectrum and making a full recovery. With a multitude of physical and behavioral tests for neurological and communication skills, diagnosing ASD is very complex, subjective, time-consuming, and expensive. We hypothesize that the use of machine learning analysis on facial features and social behavior can speed up the diagnosis of ASD without compromising real-world performance. We propose to develop a hybrid architecture using both categorical data and image data to automate traditional ASD pre-screening, which makes diagnosis a quicker and easier process. We created and tested a Logistic Regression model and a Linear Support Vector Machine for Module 1, which classifies ADOS categorical data. A Convolutional Neural Network and a DenseNet network are used for module 2, which classifies video data. Finally, we combined the best performing models, a Linear SVM and DenseNet, using three data averaging strategies. We used a standard average, weighted based on number of training data, and weighted based on the number of ASD patients in the training data to average the results, thereby increasing accuracy in clinical applications. The results we obtained support our hypothesis. Our novel architecture is able to effectively automate ASD pre-screening with a maximum weighted accuracy of 84%.

1. Introduction

In the current workplace according to a Business Town report, two-thirds of the hiring decision is based on an assessment of the applicant's EQ or emotional quotient, leaving less than half of the work opportunity relying on previous experience and technical knowledge. In this case, having impaired emotional capabilities may mean losing a vital job offer and making life significantly more difficult [1]. This issue is what people with ASD or Autism Spectrum Disorder have to cope with in a multitude of places in their everyday life which prevents them from fulfilling their full potential. 1 in every 54 children in the US suffer from some level of ASD, which accounts for 31 million people who suffer from ASD in the US alone. Worldwide, there are around 171 million people with ASD [2]. Fortunately, ASD is able to be cured as long as it is discovered, diagnosed, and treated while the patient is still at a young age, which renders the average current diagnosis time of 3 to 3.5 years unacceptable. Additionally, the cost of around $60,000 per year per patient or on average $180,000 for a complete diagnosis further prevents people from getting the diagnosis and treatment that they desperately need in a timely manner. These issues stem from the complexity and inefficiency of the current diagnosis steps, which consist of developmental monitoring, small physical tests, checklists, and an extensive full-body behavioral evaluation. The development monitoring step is made up of observations during periodic health checkups. Next, specialists administer small physical tests, such as checking for abnormal movements, head circumference, and general motor skills. Checklists, such as ADOS, ADI-R, M-CHAT, and E-2, are then used to assess a patient's ASD with a more consistent test. Lastly, specialists will complete an extensive test with an in-person evaluation. The entire process is highly subjective with a high likelihood of multiple interpretations of the same symptoms, making the current diagnosis process very inconsistent, time-consuming, and expensive [3][4][5]. This significant portion of the entire world population needs a solution that can effectively diagnose or predict if a patient has Autism in a rapid and cost-efficient way without compromising real-world performance. In this paper, we propose to utilize machine learning models in a multi-modular architecture to help automate ASD pre-screening to reduce the required time for a complete diagnosis. Pre-screening uses up a substantial amount of manpower that can be most effectively applied where their specialist training is most required. It will free up doctors' time to treat real ASD patients if machine learning models can be used to automate ASD pre-screening.

There have been many previous research works that have also improved the inefficient diagnosis of ASD using machine learning, most commonly focusing on speeding up ASD diagnosis and determining the best models and features for ASD diagnosis. The first limitation we observed from those research works was the lack of data flexibility during operation. Past machine learning solutions only analyze one form of data at a time, either checklist data or video data. Since facial feature analysis and checklist evaluation are the two of the most important factors in an ADOS diagnosis, if the model cannot consider both data, it will leave out a significant amount of potential variability as well as data flexibility for medical specialists. Another limitation is the unbalanced training and testing data that was used. Social behavior data from the ADI-R database, for example, primarily uses ASD patient data, rarely adding non-ASD patient data. This is due to more ASD patients taking the ADI-R checklist compared to non-ASD patients. This skew in the data is then reflected in the results during clinical operation [6][7][8].

## 2. Databases Used in This Research

The most common databases for ASD diagnosis from previous works were the AGRE ADOS database, ADI-R database, Simons database, ADDM database, SRS database, SSC database, and other forms of self-gathered data, whether that is from publicly available data, such as YouTube for videos, or self-created, such as home videos. There are total of 3 databases that we used in our research: the AGRE/ADOS database, Kaggle database, and a self-gathered video test dataset with corresponding ADOS data [8][9]. The proposed diagnosis methods for ASD pre-screening is based on machine learning architecture as shown in Figure 1. We first used the AGRE database of ADOS or the Autism Diagnostic Observation Schedule checklist results containing 5 and 10 classifying features for Module 1 of our hybrid architecture. The ADOS database is split into 5 modules based on age as well as verbal communication ability of the patient. This ranges from module 1, which consists of young children who are not able to use complete phrases to communicate, to module 5, which consists of teenagers and adults who are fluent in verbal communication. The ADOS evaluation is made up of 8 tasks for the patient of varying purpose and complexity. These tasks include telling a story or playing with building blocks, from which the specialists analyze a wide range of factors, including eye contact, repetition of certain words or phrases, and motor skills. These factors in the checklist are identified by an acronym, such as A1 or B3, with a score ranging from 0 to 3 and 7 to 9, measuring the severity for each factor. For example, a score of 2 means that the behavior of type specified is present and meets specific mandatory criteria [10]. In order to allow our final solution to ultimately reduce the amount of work required for pre-screening, we only analyze 5 and 10 classifying features in the two models described below.

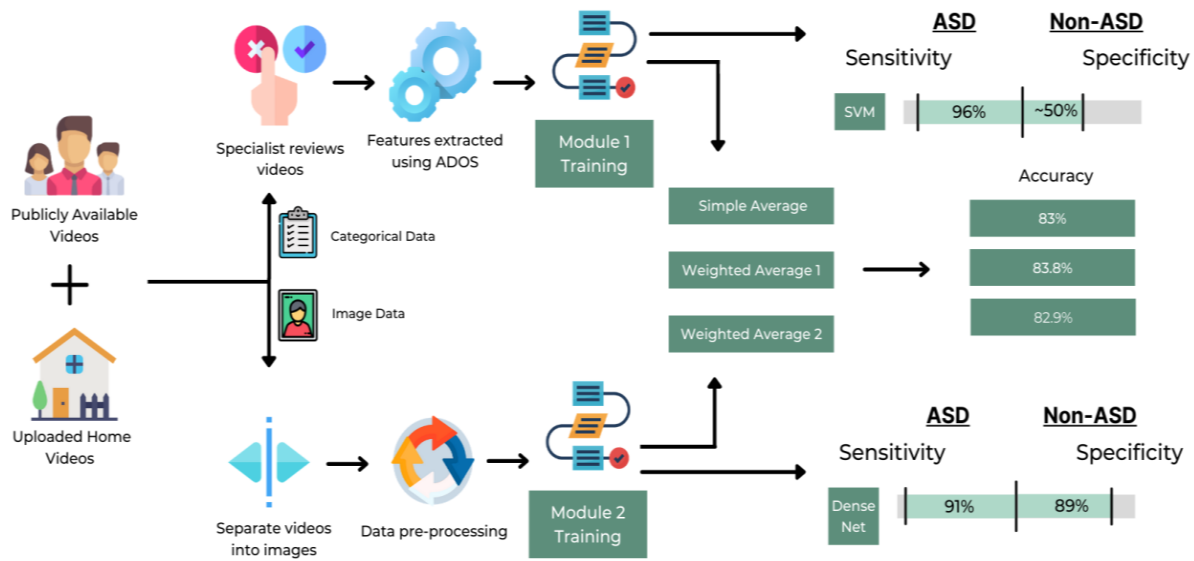

Figure 1: Proposed diagnosis method for ASD pre-screening based on novel Machine Learning architecture. This method fully utilizes two sets of clinical data: numeric social behavior data and facial image data.

The Kaggle database was used for the facial feature-based module. It contains images of ASD and non-ASD children, and the images are all within the ADOS module 2, which represents children who cannot use fluent communication. These images in ASD and non-ASD categories are split into training, testing, and validation data with 90% for training and 10% for testing. There are 1268 images for training, 150 images for testing, and 100 images for validation, in both the ASD and non-ASD categories [9]. The images are all cropped 224 x 224 x 3, focused, and mostly centered faces of the children.

In order to be able to test a hybrid architecture with two different models, classifying categorical data as well as video/image data, we needed one set of patients with both ADOS classification data and home videos to fairly and properly test the effectiveness of our architecture. After extended research, we were not able to find quality datasets that meet our criteria for our research, so we compiled the dataset ourselves. By gathering publicly available videos of both

ASD patients and non-ASD patients we were able to use around 50 videos for testing. Since we are not professional psychiatrists, we don't have the ability to qualify the videos with ADOS analysis. To help us complete this step, Dr. Hui Qi Tong from the Stanford Department of Psychiatry & Behavioral Sciences was generous enough to complete an ADOS analysis on the 50 compiled videos. After the videos were scored, we created our own database for our social behavior-based module, the Numerica Social Behavior (NSB) database, which contained the scores from the 50 videos we gathered. Also, we extracted approximately 125 images from these videos, which we input into another database for our facial feature-based module, the Videos and Image Facial Feature (VIFF) database.

## 3. Machine Learning Models

### 3.1 Social Behavior Based Module

We first selected the alternating logistic regression (ALR) for our social behavior-based module because of its ability to run very efficiently, allowing for more experimentation, despite its struggling to detect complex patterns in the data [11]. Our initial tests confirm the results from previous works, but to leverage the fast run times, we optimized the learning rate or C of the model which allowed us to gain a slight edge over previous papers. To test the limits of classifying this dataset and trying to achieve even better results, we also ran a Linear SVM since it is best suited for small to midsize databases, which is a perfect fit for our ADOS dataset of 1000 to 2500 data points depending on the number of classifying features. In the end, the Linear SVM's results were more favorable in terms of accuracy, sensitivity, and precision, so it was chosen to represent the social-behavior-based module in our hybrid architecture.

### 3.2 Facial Feature Based Module

For our facial feature-based module, we used a DenseNet model. At first, we decided to test a traditional Convolutional Neural Network (CNN) model due to its efficient runtime and its high accuracy. However, after testing, although the accuracy was very high, the model was not effectively running due to overfitting issues as well as lack of high performance in measuring sensitivity and precision. Therefore, we decided to test out a DenseNet model, a modified version of the CNN, which provided much better results. For example, the DenseNet model used a custom callback during training to adjust the learning rate and save the best weights in order to achieve optimal results. The initial results were around the same as previous works, so we decided to optimize the model by putting images extracted from the home videos we gathered into the validation dataset, improving the accuracy, sensitivity, and precision as shown in Table 1 and Figure 2. Since the validation dataset is already used to optimize the model, adding our own data will optimize the model even more [12]. In the end, the DenseNet model's results outperformed the CNN model, so the DenseNet model was chosen to represent the facial feature-based module in our hybrid architecture.

### 3.3 Proposed Hybrid Model

For our hybrid model, we integrated both NSB and VIFF datasets with their corresponding module. Since our self-gathered dataset is not extremely large, we did not use it for our social behavior and facial feature-based module tests. However, for the hybrid model, we needed a balanced, consistent, and paired dataset in order to achieve the best results. We defined three combination strategies to integrate our two results with each other: 1) standard average, 2) weighted based on the number of training data, and 3) weighted based on the number of ASD patients in the training data. We combined our two best-performing models from the two modules, the Linear SVM and the DenseNet model. We chose to run a standard average and used the results as a baseline to compare our weighted averages. Since one of the most influential factors that determine a model's accuracy is the amount of training data, we chose it to weight our two module's results. Also, a significant contributor to the final accuracy, how balanced the training dataset is also very important to consider, so we used the number of ASD patients in the training data as our second weighting strategy. This novel diagnosis method allows medical specialists to fully utilize two forms of clinical datasets, behavioral symptoms and facial features for a comprehensive ASD evaluation. It also provides doctors with the flexibility to decide the most effective way of diagnosing ASD on a case by case basis, such as specialist's availability or lack of videos. Lastly, this hybrid architecture provides the advantage of complementary results, with both modules compensating for each other's accuracy and specificity.

| References | Training & Testing Data | # ASD | # Non-ASD | Age Range | Test Accuracy | Test Sensitivity | Test Precision |
|---|---|---|---|---|---|---|---|
| Oxford CNN[13] | 2940 | 1470 | 1470 | 2-14 | 85% | 76% | 86.67% |
| DNN[14] | 3500 | 2641 | 859 | 2-14 | 70% | 74% | 63% (spec) |
| SVM[14] | 3500 | 2641 | 859 | 2-14 | 65% | 68% | 62% (spec) |
| Random Forest[14] | 3500 | 2641 | 859 | 2-14 | 63% | 69% | 58% (spec) |
| Random Forest[15] | 500 | 250 | 250 | 2-14 | 77.26% | 81.52% | 73.09% |
| Our Work: CNN | 2836 | 1418 | 1418 | 2-8 | 99% | 66% | 66% |
| Before Optimization: DenseNet | 3652 | 1828 | 1824 | 2-8 | 81% | 81% | 83% |
| After Optimization: DenseNet | 2836 | 1418 | 1418 | 2-8 | 91.88% | 91% | 89% |

Table 1: Summary and comparison with previous work for our module 2.

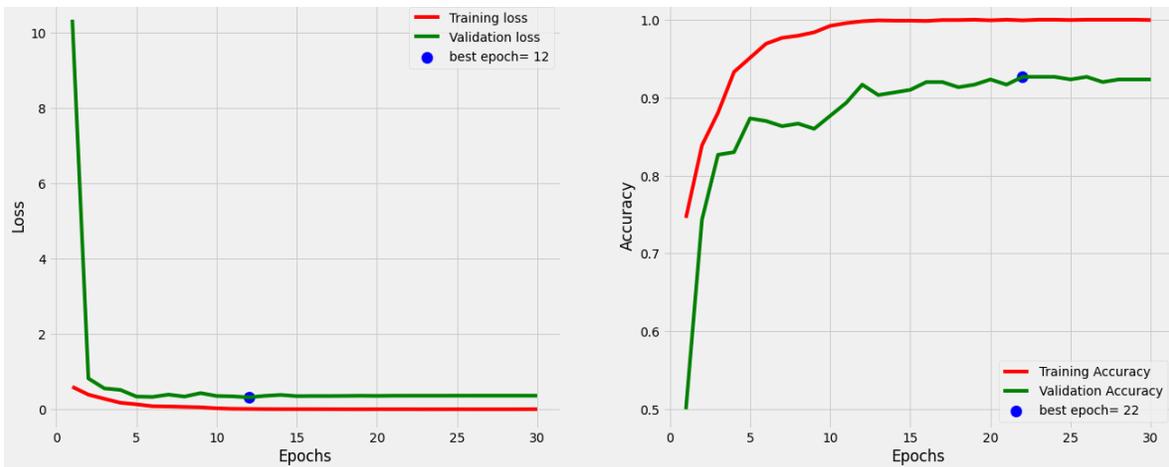

Figure 2: Graph of training and validation loss and accuracy for the facial feature-based module.

## 4. Results and Discussions

We started our model training by splitting our ADOS data into 80% and 20% pieces of the original dataset. Using 80% of the data as training and 20% of the data as testing in the social behavior module, we were able to replicate the results of at least one previous work with both the ALR and the Linear SVM for model accuracy, sensitivity, and precision [8]. However, after learning rate optimization of the model recorded in Figure 3, we were able to outperform some previous works in terms of accuracy by 2%, and both sensitivity and precision by 1% as shown in Table 2. This is because we found the most suitable learning rate for the ADOS dataset we are using. A learning rate that is too low leads to low accuracy because it does not fit the data very well. A learning rate that is too high also leads to lower accuracy during testing, especially during real-world implementation, due to overfitting. In this case the model fits the training dataset so well, that it is not able to detect an easily predicted difference.

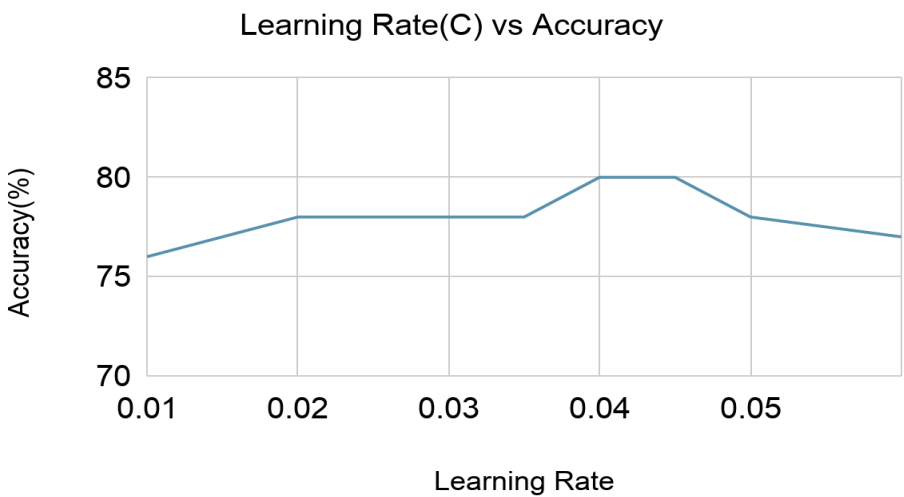

Figure 3: Graph of learning rate curve compared to accuracy for social behavior-based module optimization.

| References | Medical Record Source | # features | Training/ Testing Size | $N_{ASD}$ | $N_{Non-ASD}$ | Mean Age (SD) | % Male (N) | Test Accuracy | Test Sensitivity | Test Precision |
|---|---|---|---|---|---|---|---|---|---|---|
| Logistic Regression [8] 2018 | ADOS Module 3 | 10 | Train/Test 2870 | 2,870 | 273 | 9.08 (3.08) | 81% (N=2,557) | 76.1% | 100% | 76% |
| Logistic Regression [7] 2019 | ADOS Module 2&3 | 6/10 | Train/Test 4540 | 4,189 | 273 | ~10 (~3) | ~80% (N=~2,500) | 89.4% | 98% | 98% |
| Before Optimization Logistic Regression | ADOS Module 3 | 10 | Train/Test 2870 | 2,870 | 273 | 9.08 (3.08) | 81% (N=2,557) | 78% | 77% | 72% |
| After Optimization Logistic Regression | ADOS Module 3 | 10 | Train/Test 2870 | 2,597 | 273 | 9.08 (3.08) | 81% (N=2,557) | 80% | 78% | 73% |
| Linear SVM [8] 2018 | ADOS Module 3 | 10 | Train/Test 1,319 | 1046 | 273 | 6.92 (2.83) | 80% (N=1,101) | 73.9% | 95% | 71% |
| Linear SVM [7] 2019 | ADOS Module 2&3 | 6/10 | Train/Test 4540 | 4,189 | 343 | ~10 (~3) | ~80% (N=~2,500) | 97% | 97% | 97% |
| Before Optimization Linear SVM | ADOS Module 2 | 5 | Train/Test 1,319 | 1046 | 273 | 6.92 (2.83) | 80% (N=1,101) | 95% | 95% | 77% |
| After Optimization Linear SVM | ADOS Module 2 | 5 | Train/Test 1,319 | 1046 | 273 | 6.92 (2.83) | 80% (N=1,101) | 96% | 96% | 77% |

Table 2: Summary of prediction results from social behavior-based model.

By using the balanced Kaggle dataset, we were able to not only produce results that matched most previous works but also heavily improve on these initial results through optimization. Because the validation dataset serves as a tool for optimizing the model, inputting most of the images extracted from the videos would help optimize the DenseNet model. Also, the custom callback feature mentioned earlier adjusts the learning rate and weights every time the model trains in order to optimize the model. After optimization, we were able to outperform the previous works in terms of accuracy by 6%, sensitivity by 15%, and precision by 3%, as shown in Table 1. These optimization techniques were able to improve the results of the DenseNet model, creating a stronger analysis of facial features.

The summary of prediction results from the hybrid model after we incorporated our own dataset are shown in Table 3. We took the results from the previous two modules and used a standard average to combine the two results as well as two weighted averages that are proportional to the number of training data and the number of ASD patients in each training data. We chose to base our first weight on the number of training data because it is one of the most important factors in determining the accuracy of a model. Next, we used the number of ASD patients for our second weight because dataset bias is also an extremely influential factor in a model's performance. This combined test was an essential step in our experimentation because it represents clinical scenarios. The existing data are unbalanced since the vast majority of both datasets are ASD positive patients, only 10% for Module 1 and 32.5% for Module 2 represent non-ASD patients. After extensive research, we were not able to find a balanced dataset with the quality we required, so we decided to create our own. Through collecting home videos displaying ASD behavior among children, we were able to gather around 50 videos and 125 images that were tailored to our models and research, making facial features easily identifiable to our models. To simulate a more balanced clinical scenario, we tested our hybrid model using this combined self-created dataset. Despite having a lower accuracy compared to the facial feature-based module as seen in Table 4, our hybrid architecture compensates with a higher sensitivity. Similarly, the sensitivity of our hybrid architecture is slightly lower than that of the social behavior-based module, but it compensates with a higher accuracy. Through the balancing of the three metrics, we are able to counteract the bias from heavily skewed training data in a way that previous research has not achieved.

|  | Accuracy | Sensitivity | Precision |
| --- | --- | --- | --- |
| Module 1: Categorical data based | 80% | 80% | 73% |
| Module 2: Image data based | 86% | 74% | 83% |
| Simple Average | 83% | 78.5% | 78% |
| Weight based on # of training & testing data | 83.8% | 78.% | 79.4% |
| Weight based on # of ASD in training data | 82.9% | 78.5% | 77.9% |

Table 3: Summary of prediction results from hybrid model.

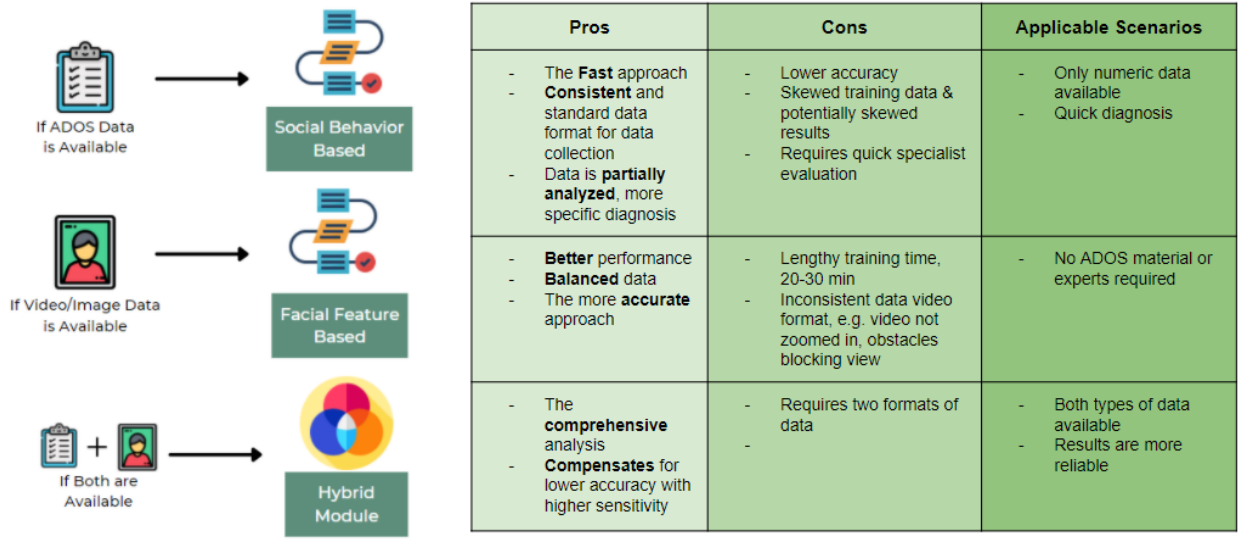

Table 4: Summary of pros and cons and data flexibility for each module.

## 5. Conclusions

Previous works show that single models are capable of achieving a certain degree of accuracy, sensitivity, and precision. But the data used in the tests are unbalanced and the model can only analyze one data format [6][7][8]. Our proposed hybrid architecture in this research is able to effectively speed up the diagnosis of ASD without compromising clinical performance. Our study is able to diagnose ASD at an effective accuracy, 84%, with balanced sensitivity and precision, while combining results from 2 separate models. This allows our solution to be applicable in a vast array of places, ranging from small rural clinics to large medical institutions. A major advantage of our novel architecture is the flexibility of required data. Since our architecture consists of two separate models, we are able to either run each model separately or in a hybrid configuration depending on the available data. For example, a patient's diagnosis is very urgent and past ADOS evaluation data is available, but a home video is not. In this circumstance, we can use a social behavior module to analyze this patient with acceptable accuracy. On the other hand, a small rural clinic may not have ASD specialists who are able to accurately evaluate a patient using ADOS, but a short home video is available. In this case our facial feature-based module can also be independently used to analyze this patient. Although these methods have acceptable performance, in an ideal situation with both video and ADOS data available, taking advantage of a hybrid architecture allows for an overall more balanced and comprehensive evaluation. With both models compensating each other for accuracy and specificity as well as being tested on a balanced self-made dataset, our hybrid architecture is able to provide more accurate results. It is also more comprehensive than past research, covering both social behavior and facial features, the two areas that are most impacted by ASD. By shrinking the time needed for a complete diagnosis with automating pre-screening, our approach can help reduce financial, psychological, and physical burdens involved when receiving an ASD diagnosis.

Moving forward, we plan to continue gathering videos of ASD and non-ASD patients to enlarge our testing database. Further testing will be conducted to determine the most efficient and viable way to score each video based on ADOS. Another feature to be implemented is the ability to identify the severity of ASD and provide the patient and the medical team with more information which can be used to determine the most suitable treatment plan. Lastly, we plan to expand our pre-screening solution to diagnose other developmental disorders, such as Parkinson's or Attention-deficit hyperactivity disorder (ADHD).

**References**

1. Understanding emotional intelligence and its importance when hiring. (2017, July 10). Retrieved March 05, 2021, from https://businesstown.com/understanding-emotional-intelligence-importance-hiring/#:~:text=Companies%20studied%20found%20that%20EQ,a%20company%20higher%20quality%20employees.
2. Autism Statistics and Facts. (n.d.). Retrieved from https://www.autismspeaks.org/autism-statistics
3. Price, S. (2019, September 27). Questions raised about accuracy of autism screening. Retrieved from https://www.healtheuropa.eu/accuracy-of-autism-screening/93640/#:~:text=Overall, 2.2% of children in,Prevention (CDC) estimates nationally
4. Testing for Autism. (2020, September 2). Retrieved March 4, 2021, from https://www.healthline.com/health/autism-tests#takeaway
5. Screening and diagnosis of autism spectrum disorder. (2020, March 13). Retrieved March 05, 2021, from https://www.cdc.gov/ncbddd/autism/screening.html
6. Wall, D., Kosmicki, J., DeLuca, T. *et al.* Use of machine learning to shorten observation-based screening and diagnosis of autism. *Transl Psychiatry* 2, e100 (2012). https://doi.org/10.1038/tp.2012.10
7. Wall DP, Dally R, Luyster R, Jung J-Y, DeLuca TF (2012) Use of Artificial Intelligence to Shorten the Behavioral Diagnosis of Autism. PLoS ONE 7(8): e43855. https://doi.org/10.1371/journal.pone.0043855
8. Tariq Q, Daniels J, Schwartz JN, Washington P, Kalantarian H, Wall DP (2018) Mobile detection of autism through machine learning on home video: A development and prospective validation study. PLoS Med 15(11): e1002705. https://doi.org/10.1371/journal.pmed.1002705
9. Gerry, Detect Autism from Facial Image, Retrieved January 12, 2021, https://www.kaggle.com/gpiosenka/autistic-children-data-set-traintestvalidate
10. McCrimmon, A., Dr. (2016, July 14). Autism spectrum Disorder: Early Detection, resilience and Growing Int... Retrieved March 10, 2021, from https://www.slideshare.net/UniversityofCalgary/autism-spectrum-disorder-early-detection-resilience-and-growing-into-adulthood-suzanne-curtin-adam-mccrimmon-david-nicholas-university-of-calgary
11. Advantages and disadvantages of logistic regression. (2020, September 02). Retrieved March 10, 2021, from https://www.geeksforgeeks.org/advantages-and-disadvantages-of-logistic-regression/
12. What is The Difference Between Test and Validation Datasets? (14 July, 2017). Retrieved January 3, 2021, from https://machinelearningmastery.com/difference-test-validation-datasets/

13. Musser, Mikian. "Detecting Autism Spectrum Disorder in Children With Computer Vision." *Medium*, Towards Data Science, 24 Aug. 2020, towardsdatascience.com/detecting-autism-spectrum-disorder-in-children-with-computer-vision-8abd7fc9b40a.
14. Heinsfeld, A. S., Franco, A. R., Craddock, R. C., Buchweitz, A., & Meneguzzi, F. (2018). Identification of autism spectrum disorder using deep learning and the abide dataset. *NeuroImage: Clinical, 17*, 16-23. doi:10.1016/j.nicl.2017.08.017
15. Omar, K. S., Mondal, P., Khan, N. S., Rizvi, M. R., & Islam, M. N. (2019). A machine learning approach to predict autism spectrum disorder. *2019 International Conference on Electrical, Computer and Communication Engineering (ECCE)*. doi:10.1109/ecace.2019.8679454